\documentclass[11pt]{article}

\usepackage[final]{acl}

\usepackage{times}
\usepackage{latexsym}
\usepackage{makecell}
\usepackage{booktabs}
\usepackage{enumitem}
\usepackage{float} 
\usepackage[table]{xcolor}
\usepackage{letterspace}
\usepackage{listings}
\usepackage[most]{tcolorbox}
\usepackage{stfloats} 
\usepackage{lineno}

\usepackage{booktabs}    % For nice tables
\usepackage{tikz}        % Base TikZ library
\usepackage{pgf-pie}     % For pie charts
\usepackage{hyperref}    % For clickable links

\definecolor{turquoise}{RGB}{64, 224, 208}

\usepackage[T1]{fontenc}
\usepackage[utf8]{inputenc}

\usepackage{microtype}

\usepackage{inconsolata}

\usepackage{graphicx}

\title{Luth: Efficient French Specialization for Small Language Models and Cross-Lingual Transfer}

\author{
Maxence Lasbordes \thanks{Equal contribution.} \\
LightOn, Paris \\
Inria Paris \\
\texttt{maxence.lasbordes@lighton.ai}
\And
Sinoué Gad \footnotemark[1] \\
École Polytechnique\\
Institut Polytechnique de Paris\\
\texttt{sinoue.gad@polytechnique.edu}
}

\begin{document}
\maketitle
\begin{abstract}
The landscape of Large Language Models remains predominantly English-centric, resulting in a significant performance gap for other major languages, such as French, especially in the context of Small Language Models (SLMs). Existing multilingual models demonstrate considerably lower performance in French compared to English, and research on efficient adaptation methods for French remains limited. To address this, we introduce \textbf{Luth}, a family of French-specialized SLMs: through targeted post-training on curated, high-quality French data, our models outperform all open-source counterparts of comparable size on multiple French benchmarks while retaining their original English capabilities. We further show that strategic model merging enhances performance in both languages, establishing Luth as a new state of the art for French SLMs and a robust baseline for future French-language research.
\end{abstract}

\section{Introduction}

Large Language Models (LLMs) have shown great potential in complex multilingual tasks \citep{Grattafiori2024_llama3, OpenAI2023GPT4, Yang2025}, but performance is uneven across languages. Due to abundant English data, most research focuses on English, leaving other languages behind \citep{Ruder2022SquareOneBias, Li2024QuantifyingMultilingual}. French, spoken by over 280 million people, remains underrepresented in datasets and models, resulting in weaker performance within state-of-the-art multilingual systems.

In parallel, SLMs have emerged as a promising direction. Studies show that smaller models, when properly trained or adapted, can achieve competitive performance across diverse tasks \citep{Lepagnol2024SLMs, Nguyen2024Survey}. Their compact size enables faster inference, lower computational overhead, and practical deployment, making them well-suited for real-world applications \citep{belcak2025slm}. SLMs can also be efficiently specialized to specific languages or domains, offering a practical path to high-quality French language models without relying on large-scale resources. \\
% \noindent \textcolor{red}{Pas convaincu par le contenu de cette partie.}

% Challenges remain for multilingual SLMs: (1) scarce high-quality datasets in lower resource languages; (2) high training/adaptation costs from scratch; (3) lack of standardized benchmarks. Our work focuses on French-specific resources and adaptation strategies, contributing to a stronger open-source LLM ecosystem for underrepresented languages.

In this paper, we introduce \textbf{Luth}\footnote{\url{https://github.com/kurakurai/Luth}}, a family of compact French SLMs designed to address the English-centric bias through targeted adaptation. We demonstrate that using carefully curated post-training data, it is possible to significantly improve French capabilities, including general knowledge, instruction-following, and mathematical reasoning, without degrading original English performance, and even enhancing both languages through strategic model merging.\smallskip

\noindent Specifically, our contributions are:
\begin{enumerate}[leftmargin=1.5em, noitemsep, topsep=0pt]
    \item The \textbf{Luth-SFT}\footnote{\url{https://huggingface.co/datasets/kurakurai/luth-sft}} dataset, containing \(570\)k samples of French instruction-response pairs, which substantially improves model performance in general knowledge, instruction following, and mathematical reasoning.
    \item The \textbf{Luth}\footnote{\url{https://huggingface.co/collections/kurakurai/luth-models-68d1645498905a2091887a71}} family, including 5 models ranging from 350M to 1.7B parameters, achieving state-of-the-art performance in French within their size categories and delivering an absolute average improvement of up to $+11.26\%$ across six French benchmarks.
    \item An efficient and reproducible methodology for language-specific adaptation, easily extendable to other languages, while preserving performance in other languages.
\end{enumerate}

\begin{figure*}[!t]
    \centering
    \includegraphics[width=\textwidth]{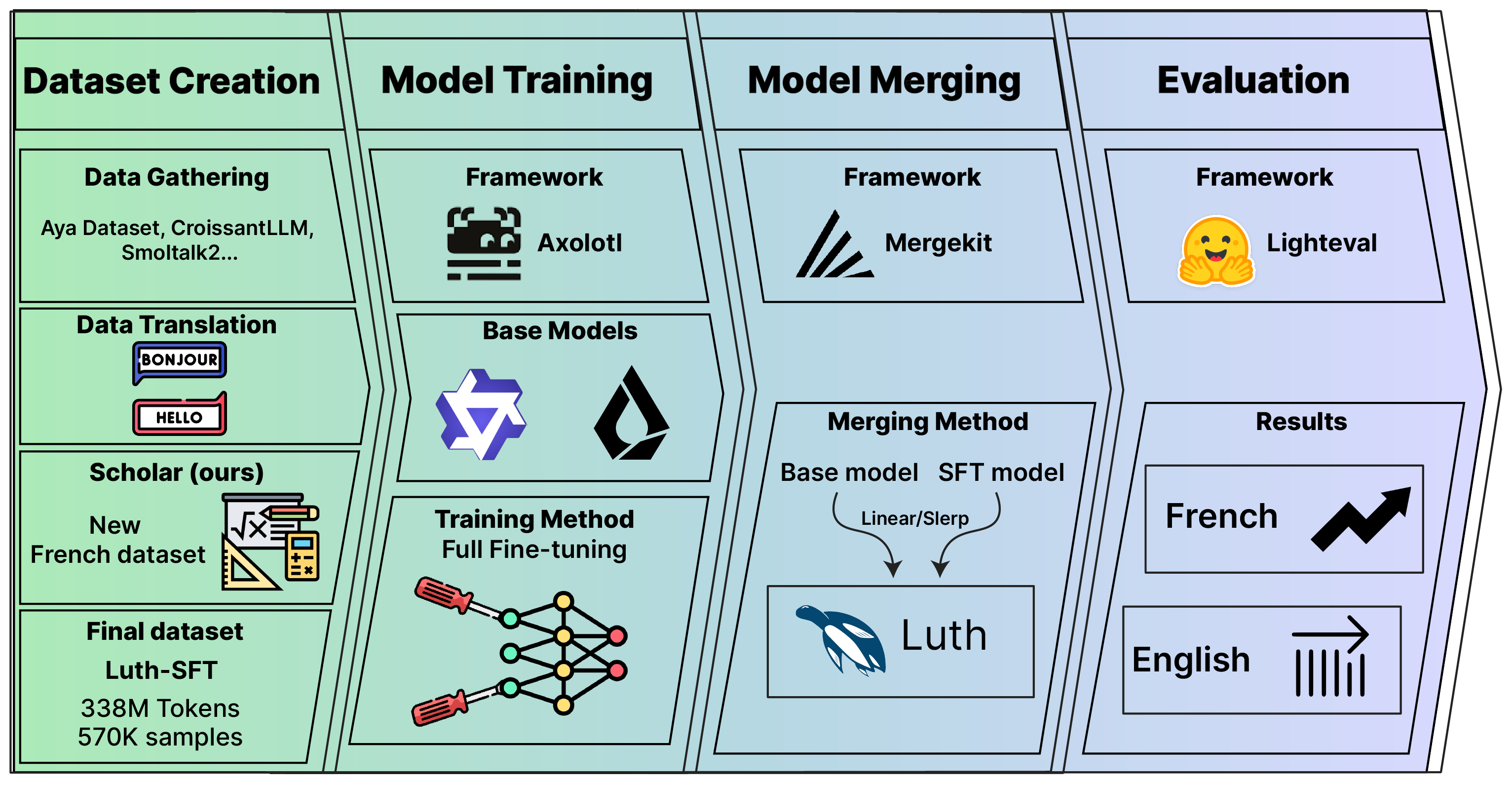}
    \caption{Overview of the four main stages in constructing the Luth models, including their substeps, methods, and frameworks.}
    \label{fig:overall_pipeline}
\end{figure*}

\section{Related Work}

The development of multilingual and language-specific models aims to mitigate the English-centric bias of current LLMs. Models such as BLOOM~\cite{LeScao2022BLOOM}, Llama~\cite{Grattafiori2024_llama3}, and AYA~\cite{Ustun2024} cover dozens of under-represented languages, but they do not focus on language-specific optimization and often underperform on individual languages. Regional initiatives, such as EuroLLM~\cite{Martins2024EuroLLM} and Apertus~\cite{apertus2025apertusdemocratizingopencompliant}, aim to improve multilingual coverage, with Apertus supporting over 1,000 languages and emphasizing data compliance.

Several efforts focus specifically on French. Early work includes PAGnol~\cite{launay2021pagnolextralargefrenchgenerative}, which introduced scaling laws for French and trained a 1.5B-parameter GPT model. More recent contributions include CroissantLLM~\cite{Faysse2024}, a French–English bilingual model; Gaperon~\cite{godey2025gaperonpepperedenglishfrenchgenerative}, a fully open suite of French–English–code models emphasizing transparency and reproducibility; Lucie~\cite{Gouvert2025Lucie}, which open-sourced substantial resources for French LLM development; and Pensez~\cite{Ha2025}, which studied French models with a focus on reasoning and data quality.

Despite these contributions, important gaps remain. Many works prioritize large, resource-intensive models or report performance shortfalls relative to multilingual baselines of comparable size. Moreover, they offer few practical, low-cost recipes to substantially improve French-language capabilities, leaving room for compact, French-specialized models and efficient adaptation strategies suitable for resource-constrained settings.

\section{Luth-SFT Dataset}

\begin{figure*}[!t]
    \centering
    \includegraphics[width=\textwidth]{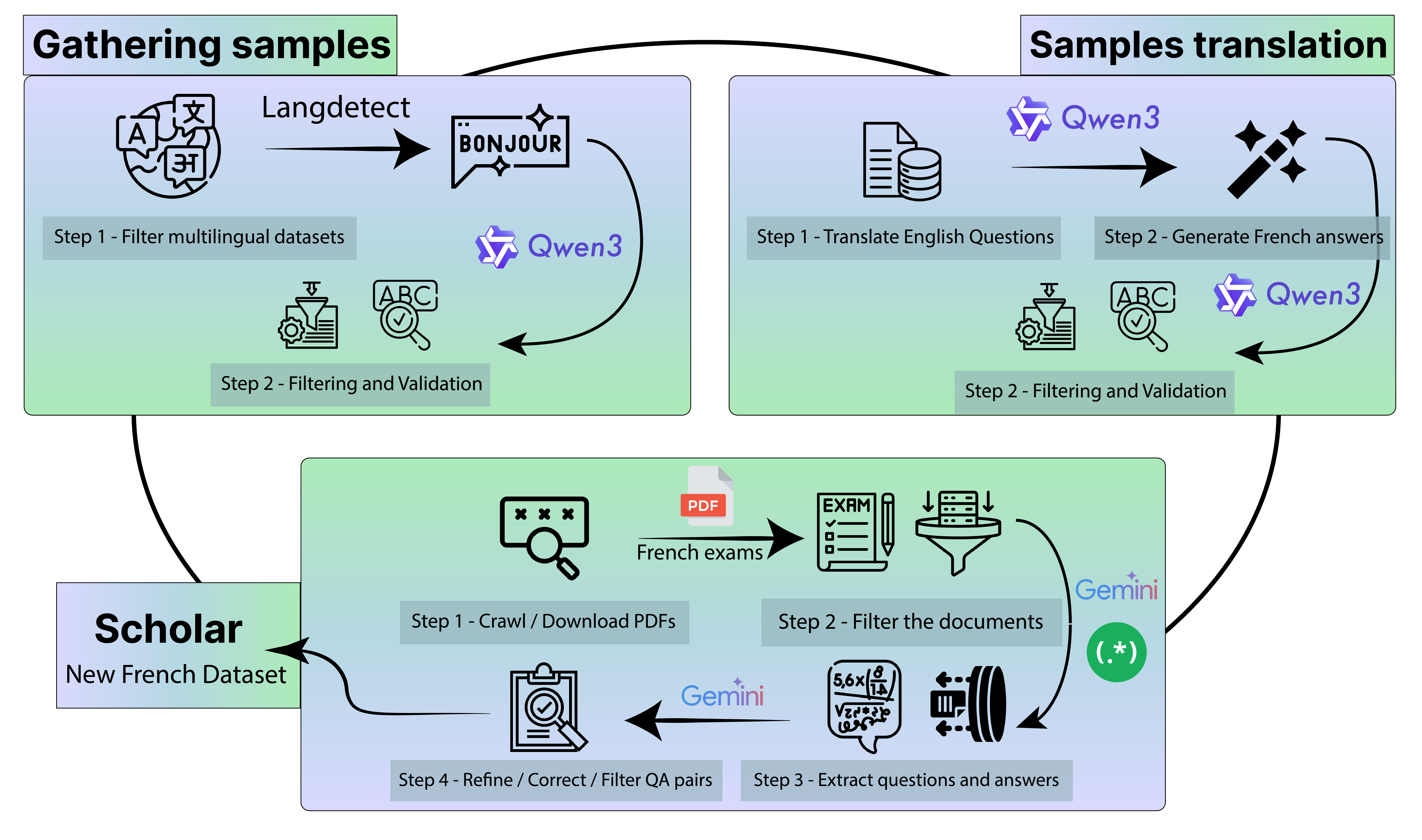}
    \caption{Overview of the Luth-SFT dataset construction pipeline, from data collection and translation to filtering and the Scholar subset creation.}
    \label{fig:dataset_pipeline}
\end{figure*}

To address the lack of high-quality open-source French post-training datasets, we introduce \textbf{Luth-SFT}, which contains \(570\)k samples (\(338\) million tokens) of French instruction–response pairs (Figure~\ref{fig:dataset_pipeline}).\smallskip

\noindent\textbf{Data Gathering} \quad To build this dataset, we first collected parts from existing multilingual datasets, including AYA~\cite{Ustun2024}, Smoltalk2~\cite{huggingfacetb2025smoltalk2}, and CroissantLLM~\cite{Faysse2024}. As the datasets are massively multilingual, we language filtered the French samples via the langdetect libary~\cite{langdetect}. \smallskip

\noindent\textbf{Data Translation} \quad To further diversify and expand our French dataset, we selected two high-quality, openly available English instruction datasets, Tülu 3~\cite{Lambert2024} and OpenHermes~\cite{OpenHermes}. Our approach is twofold: (1) translate the English prompts into French (\ref{sec:appendixTranslation}) using strong multilingual models (GPT-4o and Qwen3 32B in non-reasoning mode), and (2) generate new French responses from scratch conditioned on the translated prompts, rather than directly translating the original answers. For Tülu 3, we focused exclusively on the math and instruction-following subsets, as these align with our objectives. The samples produced through this pipeline constitute the majority of our dataset. Notably, for OpenHermes, an existing French version generated with GPT-4o following this methodology was already available, substantially reducing the associated computational cost~\cite{alhajar2025openhermesfr}. \smallskip

\noindent\textbf{Filtering} \quad We used a two-stage filtering pipeline to ensure both dataset quality and domain relevance. The first stage, \underline{linguistic validation}, enforces strict French language criteria, including grammatical correctness, coherence, absence of code-switching or mixed-language content, and proper instructional formatting. The second stage, \underline{content filtering}, systematically removes samples from three categories: programming-related content (e.g., code snippets, debugging queries, tool discussions), tool-calling content (e.g., API usage, command-line operations, system configuration), and samples containing logical inconsistencies or factual errors. This approach preserved instruction-following, scientific discourse, and general conversational samples while maintaining high linguistic and content standards. All system prompts used are listed in~\ref{sec:appendixFiltering}. \smallskip

\noindent\textbf{Scholar} \quad This subset was developed to address the scarcity of high-quality scientific resources in French. The dataset draws extensively from \emph{Baccalauréat} and \emph{Classes Préparatoires aux Grandes Écoles} (CPGE) examination materials, providing both questions and detailed solutions (see example snippet in \ref{sec:appendixScholar}) across a broad range of subjects. A key objective was to build a resource that is non-synthetic and rooted in expert knowledge. Examination materials were particularly well-suited for this purpose, as they are typically accompanied by official solutions in PDF format, authored and validated by domain experts. In total, more than \(\mathbf{14{,}000}\,\text{PDFs}\) were collected, covering examination sessions from 1980 to 2025\footnote{Mainly sourced from \href{https://www.sujetdebac.fr}{Sujet Bac} and \href{https://prepas.org/index.php?module=Sujets}{UPS Sujet}.}. These documents were processed through a multi-step pipeline (prompts listed in Appendix~\ref{sec:appendixA}):
\begin{enumerate}[noitemsep, topsep=0pt, leftmargin=*]
    \item Crawling and downloading the examination PDFs.
    \item Filtering the documents (some PDFs contained scanned solutions and were therefore unusable).
    \item Extracting questions and answers using a combination of regular expressions and LLM-assisted parsing with Gemini 2.5 Flash (\citealp{comanici2025gemini}).
    \item Refining LaTeX formatting for equations and enriching the solutions with additional explanatory details (\ref{sec:appendixScholar}) using Gemini 2.5 Pro (\citealp{comanici2025gemini}), as some official corrections were rather concise.
    \item Performing a final filtering step to remove anomalous samples, including misaligned questions and answers, missing data, and formatting errors.
\end{enumerate}
\smallskip

After processing, the dataset contains \(\mathbf{30{,}300~\text{samples}}\), distributed across several domains. The subject distribution is summarized Table~\ref{tab:distribution}.
It should be noted that the proportions mainly reflect the availability of data for each subject, and do not represent a deliberate choice on our part.

\begin{table}[h]
\centering
\begin{tabular}{@{}p{0.65\columnwidth} r@{}}
\toprule
\textbf{Subject} & \textbf{Percentage} \\
\midrule
Mathematics & 67.23\% \\
Physics--Chemistry & 10.61\% \\
Computer Science & 9.08\% \\
Engineering Science & 6.04\% \\
Biology & 5.51\% \\
Other (Economics, Accounting, Social Sciences) & 1.52\% \\
\bottomrule
\end{tabular}
\caption{Distribution of scholars by subject.}
\label{tab:distribution}
\end{table}

\section{Luth Models}

\subsection{Model Training}

As this work focuses on SLMs with fewer than 2B parameters, we conducted comprehensive evaluations of multilingual models in this size range to identify the best-performing model for French and to enhance its capabilities. We considered LFM2 (350M, 700M, and 1.2B)~\cite{liquidai_lfm2} and Qwen3 (0.6B and 1.7B)~\cite{Yang2025} for their strong French and English performance. While other SLMs, such as LLaMA 3.2 (1B)~\cite{Grattafiori2024_llama3}, SmolLM2 (360M and 1.7B)~\cite{benallal2025smollm2}, and Qwen2.5 (0.5B and 1.5B)~\cite{Yang2024}, are also viable alternatives, our evaluations indicate that they underperform relative to more recent models on the tasks considered in this work. The models were selected based on their capabilities in Math, General Knowledge and Instruction Following in both French and English. Qwen3 and LFM2 variants then went through a full fine-tuning stage, instead of LoRA~\cite{hu2021lora} for better learning, on our \textbf{Luth-SFT} dataset, which infuses them with a richer understanding of French, specific vocabulary, domain-specific terminology, and improved their skills in the previously mentioned areas. \\

\begin{figure}[t]
    \centering
    \includegraphics[width=\columnwidth]{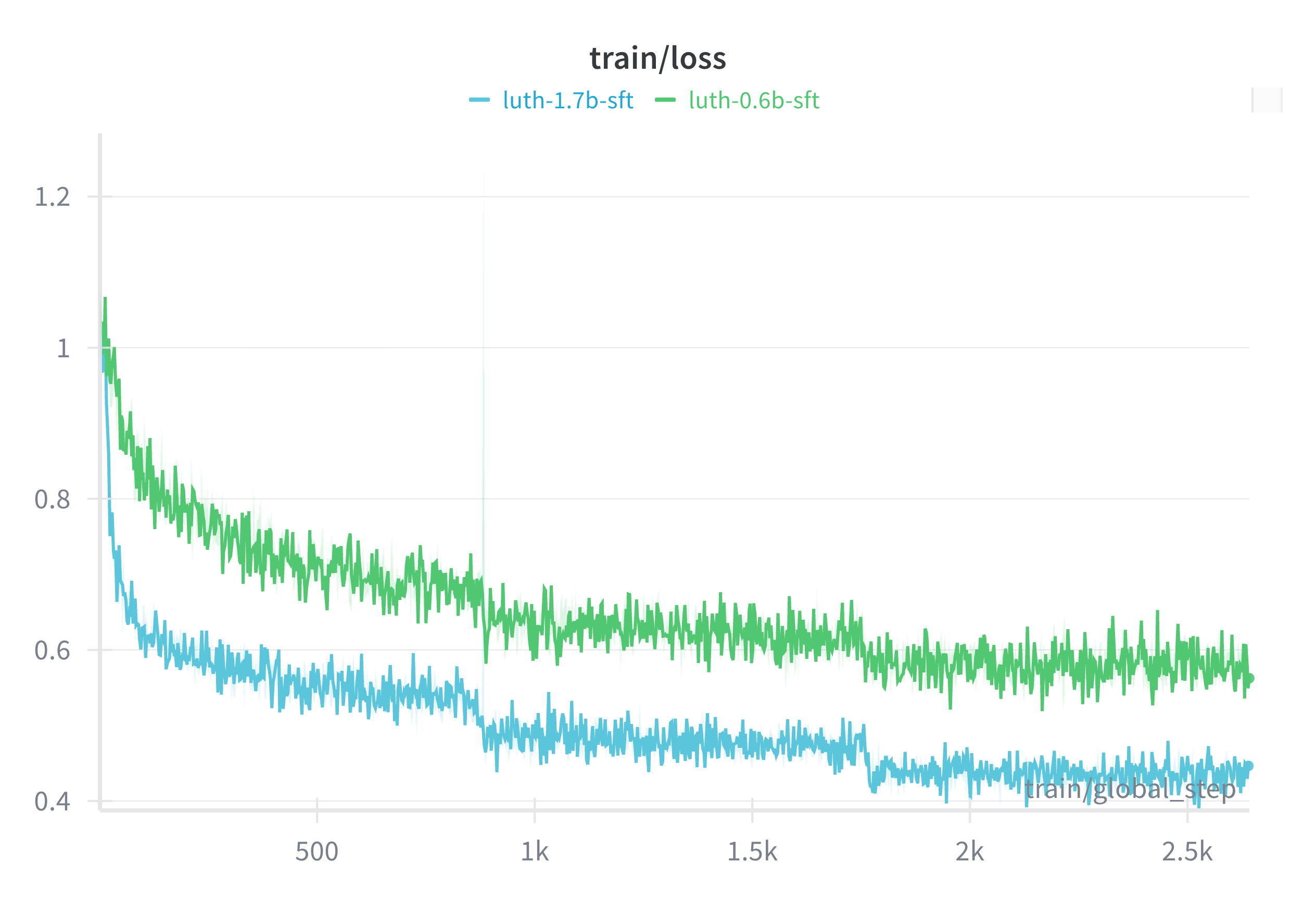}
    \caption{Loss per step during full fine-tuning on the Luth-SFT dataset over 3 epochs for Qwen3-0.6B (green) and Qwen3-1.7B (blue).}
    \label{fig:train_loss}
\end{figure}

\noindent\textbf{Full Fine-tuning} \quad We fine-tuned the models on our curated \textbf{Luth-SFT} dataset using the Axolotl framework~\cite{axolotl}. The trainings were conducted on a single NVIDIA H100 GPU (80GB VRAM) for three epochs. We used various training hyperparameters for the models, which can be found in the Appendix~\ref{sec:appendixB}. For all models, we employed FlashAttention~\cite{dao2022flashattention} to reduce memory consumption and accelerate training through memory-efficient attention computation, and sequence packing to maximize GPU utilization by concatenating multiple shorter sequences into fixed-length batches, with a maximum sequence length of $16{,}384$. For instance, Luth-0.6B-Instruct was trained with widely used hyperparameters, including a learning rate of $2 \times 10^{-5}$, an effective batch size of 24 (achieved via gradient accumulation), and a cosine learning rate scheduler with a \(10\%\) warm-up period. Examples of training losses are shown in Figure~\ref{fig:train_loss}. Due to computational limitations, we did not perform extensive hyperparameter sweeps for all models, and we leave this investigation to future work.

\subsection{Model Merging}

\begin{figure*}[!t]
    \centering
    \includegraphics[width=\textwidth]{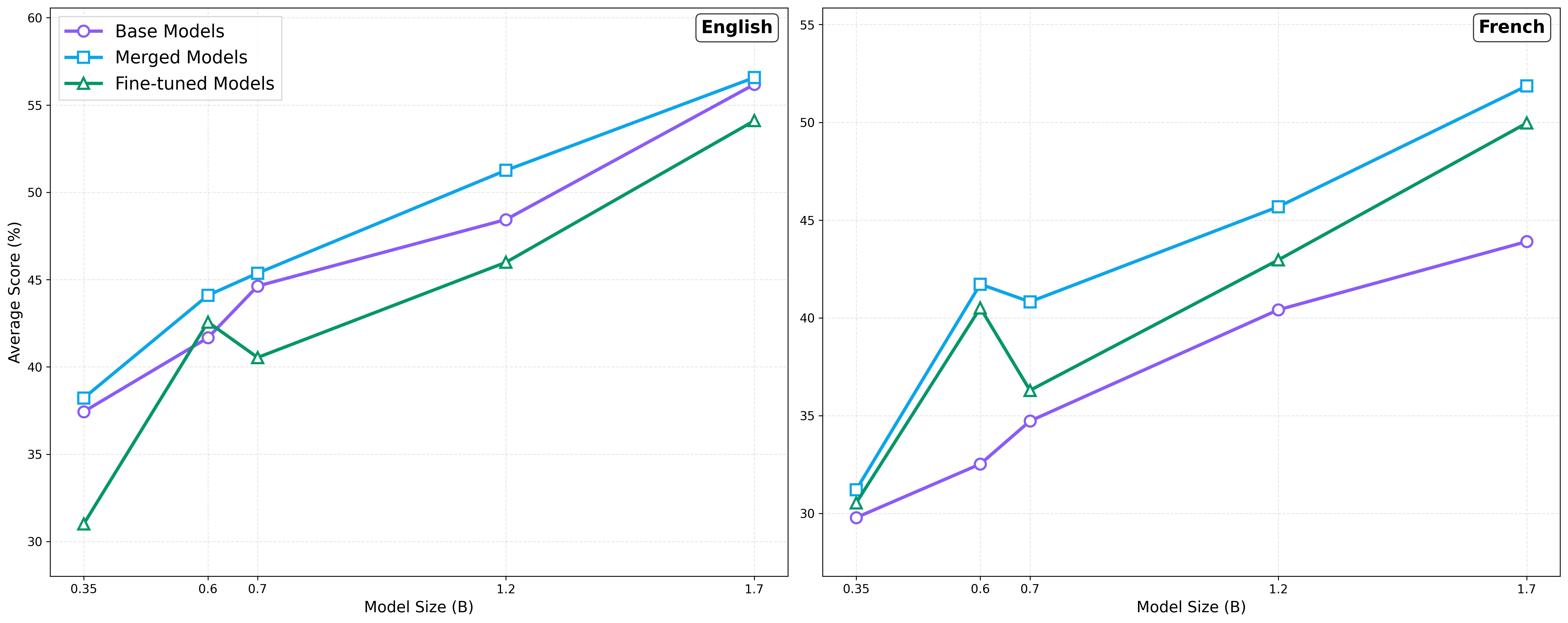}
    \caption{Performance comparison of the Luth models in their base form (e.g., Qwen3-0.6B), after fine-tuning (e.g., Qwen3-0.6B fine-tuned), and after merging (e.g., Luth-0.6B-Instruct), averaged over four French/English benchmarks: IFEval, MMLU, GPQA-Diamond, and Math500. \underline{Left panel shows English} performance, \underline{right panel shows French} performance.}
    \label{fig:en_fr_merge}
\end{figure*}

Model merging has recently gained attention as an effective technique for combining the parameters of multiple models, typically fine-tuned on different tasks or datasets, into a single system. This approach enables the merged model to inherit complementary strengths without additional retraining. Prior work has shown that merging can even outperform the individual components being merged \citep{Yang2024_merge}, a finding we confirm in our experiments (Figure~\ref{fig:en_fr_merge}).

In our setting, this method is particularly relevant: since our dataset is exclusively French, fine-tuning strongly improves French capabilities but can slightly degrade performance in other languages, including English (Figure~\ref{fig:en_fr_merge}). Model merging offers a cost-effective solution to this problem, allowing us to preserve cross-lingual abilities while still gaining improvements in French. Indeed, we observe that merging not only recovers lost English performance but also improves overall results across both languages. Moreover, merging provides a natural way to mitigate catastrophic forgetting \citep{alexandrov2024bam}.

\begin{table}[h!]
\small
  \centering
  \resizebox{1\linewidth}{!}{%
    \begin{tabular}{lccc}
      \toprule
      \textbf{Model name} & \textbf{\makecell{Base\\model}} 
                           & \textbf{\makecell{Merging\\method}} 
                           & \textbf{Coeff.} \\
      \midrule
      Luth-0.6B-Instruct & Qwen3 & SLERP & 0.7 \\
      Luth-1.7B-Instruct & Qwen3 & SLERP & 0.5 \\
      Luth-LFM2-350M & LFM2 & Linear & 0.3 \\
      Luth-LFM2-700M & LFM2 & Linear & 0.4 \\
      Luth-LFM2-1.2B & LFM2 & Linear & 0.5 \\
      \bottomrule
    \end{tabular}%
  }
    \caption{\label{tab:merging-params}
    Overview of the Luth models and key merging details that produced the most stable performance across both French and English in our experiments. The coefficient (Coeff.) indicates the proportion of the fine-tuned model used in the merge with the base model (e.g., 0.7 corresponds to 70\% of the fine-tuned model and 30\% of the base model).
  }
\end{table}

We used MergeKit, a framework that facilitates model fusion and provides a range of merging methods \citep{Goddard2024}. Since no single merging technique appears to be universally superior \citep{Yang2024_merge}, we experimented with various approaches. Surprisingly, the most stable results in our experiments were obtained with relatively simple methods, namely linear interpolation (LERP) and spherical linear interpolation (SLERP).

LERP combines two models in a straightforward linear fashion according to a coefficient $\alpha$:
\[
w = (1-\alpha) w_0 + \alpha w_1
\]

SLERP, in contrast, interpolates along the arc of the unit sphere :
\[
w = \frac{\sin((1-\alpha)\theta)}{\sin(\theta)} w_0 + \frac{\sin(\alpha \theta)}{\sin(\theta)} w_1
\]
with $\theta = \arccos(w_0 \cdot w_1)$, the angle between the two weights.

The main difference is that LERP follows a straight line in weight space, whereas SLERP follows a spherical arc, which can better preserve properties when the models are further apart. 

We therefore empirically evaluated these methods and hyperparameters, and selected the ones that provided the best results, reported in Table~\ref{tab:merging-params}.

\section{Evaluation}

% ---------------------------
% French benchmarks with 'Vous êtes un assistant utile'
% ---------------------------
\begin{table*}[!t]
\small
\centering
\begin{tabular}{lcccccc}
\toprule
\textbf{Model} & \makecell{\textbf{IFEval} \\ French} & \makecell{\textbf{GPQA-Diamond} \\ French} & \makecell{\textbf{MMLU} \\ French} & \makecell{\textbf{Math500} \\ French} & \makecell{\textbf{Arc-Challenge} \\ French} & \makecell{\textbf{Hellaswag} \\ French} \\
\midrule
\rowcolor{blue!15} Luth-1.7B-Instruct & \underline{58.53} & \textbf{36.55} & \textbf{49.75} & \textbf{62.60} & 35.16 & 31.88 \\
\rowcolor{blue!15} Luth-LFM2-1.2B & \textbf{59.95} & 28.93 & \underline{48.02} & 45.80 & \underline{38.98} & \underline{36.81} \\
Qwen3-1.7B & 54.71 & \underline{31.98} & 28.49 & \underline{60.40} & 33.28 & 24.86 \\
SmolLM2-1.7B-Instruct & 30.93 & 20.30 & 33.73 & 10.20 & 28.57 & \textbf{49.58} \\
Qwen2.5-1.5B-Instruct & 31.30 & 27.41 & 46.25 & 33.20 & 32.68 & 34.33 \\
LFM2-1.2B & 54.41 & 22.84 & 47.59 & 36.80 & \textbf{39.44} & 33.05 \\
\addlinespace[0.3em]
\midrule
\addlinespace[0.3em]
\rowcolor{blue!15} Luth-LFM2-700M & \textbf{50.22} & \underline{27.92} & \textbf{44.72} & \underline{38.40} & \textbf{36.70} & \underline{48.25} \\
\rowcolor{blue!15} Luth-0.6B-Instruct & \underline{48.24} & \textbf{34.52} & 40.12 & \textbf{44.00} & 33.88 & 45.58 \\
Llama-3.2-1B & 27.79 & 25.38 & 25.49 & 15.80 & 29.34 & 25.09 \\
LFM2-700M & 41.96 & 20.81 & \underline{43.70} & 32.40 & \underline{36.27} & 41.51 \\
Qwen3-0.6B & 44.86 & 26.90 & 27.13 & 29.20 & 31.57 & 25.10 \\
Qwen2.5-0.5B-Instruct & 22.00 & 25.89 & 35.04 & 12.00 & 28.23 & \textbf{51.45} \\
\addlinespace[0.3em]
\midrule
\addlinespace[0.3em]
\rowcolor{blue!15} Luth-LFM2-350M & \textbf{38.26} & 26.40 & \textbf{39.15} & \textbf{23.00} & \textbf{34.13} & \textbf{43.39} \\
SmolLM2-360M-Instruct & 21.50 & \underline{28.43} & 26.14 & 3.20 & 26.60 & 32.94 \\
LFM2-350M & \underline{31.55} & \textbf{28.93} & \underline{38.63} & \underline{18.00} & \underline{33.36} & \underline{39.13} \\
\bottomrule
\end{tabular}
\caption{\label{tab:french-benches}
Results of Luth and other models on various French tasks. The scores are reported as percentages (Pass@1), averaged over three runs. The highest and second-best scores are shown in \textbf{bold} and \underline{underlined} respectively for each model category.
}
\end{table*}

As the models we test were trained on a large part of English data, we also evaluate on English to assess our model's capabillities on that language after having been optimized in French with our techniques. Our evaluation process is fully transparent, and all reported results are reproducible using open-source code\footnote{\url{https://github.com/kurakurai/Luth}} and publicly available data.

\subsection{Benchmark Selection}

As mentioned in the previous sections, we focused on specific capabilities in our training data, particularly instruction following, general knowledge, and mathematics. Among the dozens of English benchmarks available, we selected widely used ones that cover these specific capabilities. For French, we relied on benchmarks from multilingual efforts or on translated versions of their English counterparts, all openly available on Hugging Face. We used six benchmarks, each available in both French and English.

\noindent\textbf{IFEval} \quad IFEval~\cite{Zhou2023IFEval} is a benchmark designed to evaluate instruction following and alignment abilities of language models, testing how well they adhere to and execute given instructions across diverse contexts.

\noindent\textbf{Math500} \quad Math~\cite{Hendrycks2021MATH} is a mathematical reasoning dataset containing 500 problems ranging from arithmetic to higher-level mathematics, assessing models’ problem-solving and reasoning skills.

\noindent\textbf{GPQA-Diamond} \quad GPQA~\cite{Rein2023GPQA} focuses on general knowledge question answering, providing challenging multiple-choice questions to test factual and commonsense reasoning.

\noindent\textbf{MMLU} \quad MMLU~\cite{Hendrycks2021MMLU} is a broad benchmark covering 57 subjects, including humanities and STEM, designed to evaluate general knowledge and multitask understanding.

\noindent\textbf{Arc-Challenge} \quad The AI2 reasoning challenge dataset~\cite{Clark2018ARC} consists of difficult multiple-choice science questions aimed at testing reasoning skills in grade-school science topics.

\noindent\textbf{HellaSwag} \quad HellaSwag~\cite{Zellers2019HellaSwag} is a commonsense reasoning benchmark that requires models to select the most plausible continuation of a story or scenario, emphasizing context-dependent understanding.

\subsection{Evaluation workflow and Reasonning mode}

Most available evaluation frameworks provide limited support for French benchmarks, as they focus predominantly on English and offer minimal coverage of multilingual tasks. We chose to use LightEval~\cite{HuggingFace2024} due to its simplicity and its ability to easily add custom tasks. We added all the benchmarks mentioned above to our setup, along with their corresponding prompts and metrics in French. 

The latest version of LightEval did not provide a mechanism to toggle reasoning mode for hybrid models. We modified it to add an \texttt{enable\_thinking} option, allowing explicit control over the inclusion of reasoning traces enclosed in \texttt{<think></think>}. This extension was particularly important for Qwen3, which defaults to reasoning mode, as it enabled us to conduct all evaluations in non-reasoning mode.

We also extended LightEval to allow toggling \texttt{enable\_prefix\_caching} to \texttt{false}, since this feature is not supported by LFM2 models. Finally, we adapted the latest version of vLLM (0.10.2) to ensure compatibility with LightEval.

\subsection{Results}
We present the results of our five Luth models against several strong multilingual SLMs in Tables~\ref{tab:french-benches} and~\ref{tab:english-benches}, for French and English respectively. Scores for each benchmark were computed as the average of three runs (\( \text{temperature} = 0 \)), using the same system prompts --- \texttt{"You are a helpful assistant."} for English and \texttt{"Vous êtes un assistant utile."} for French.
 \\

 % ---------------------------
% English benchmarks (models as rows, benches as columns)
% ---------------------------
\begin{table*}[!t]
\small
\centering
\begin{tabular}{lcccccc}
\toprule
\textbf{Model} & \makecell{\textbf{IFEval} \\ English} & \makecell{\textbf{GPQA-Diamond} \\ English} & \makecell{\textbf{MMLU} \\ English} & \makecell{\textbf{Math500} \\ English} & \makecell{\textbf{Arc-Challenge} \\ English} & \makecell{\textbf{Hellaswag} \\ English} \\
\midrule
\rowcolor{turquoise!30} Luth-1.7B-Instruct & 65.80 & 29.80 & \textbf{60.28} & \underline{70.40} & 42.24 & 58.53 \\
\rowcolor{turquoise!30} Luth-LFM2-1.2B & \textbf{70.55} & \underline{30.30} & 54.58 & 50.60 & \textbf{43.26} & 58.42 \\
Qwen3-1.7B & \underline{68.88} & \textbf{31.82} & 52.82 & \textbf{71.20} & 36.18 & 46.98 \\
SmolLM2-1.7B-Instruct & 49.04 & 25.08 & 50.27 & 22.67 & 42.32 & \textbf{66.94} \\
Qwen2.5-1.5B-Instruct & 39.99 & 25.76 & \underline{59.81} & 57.20 & 41.04 & \underline{64.48} \\
LFM2-1.2B & 68.52 & 24.24 & 55.22 & 45.80 & \underline{42.58} & 57.61 \\
\addlinespace[0.3em]
\midrule
\addlinespace[0.3em]
\rowcolor{turquoise!30} Luth-LFM2-700M & \underline{63.40} & 29.29 & \underline{50.39} & 38.40 & \textbf{38.91} & \underline{54.05} \\
\rowcolor{turquoise!30} Luth-0.6B-Instruct & 53.73 & 25.76 & 48.12 & \textbf{48.80} & 36.09 & 47.03 \\
Llama-3.2-1B & 44.05 & 25.25 & 31.02 & 26.40 & 34.30 & \textbf{55.84} \\
LFM2-700M & \textbf{65.06} & \textbf{30.81} & \textbf{50.65} & 32.00 & \underline{38.65} & 52.54 \\
Qwen3-0.6B & 57.18 & \underline{29.29} & 36.79 & \underline{43.40} & 33.70 & 42.92 \\
Qwen2.5-0.5B-Instruct & 29.70 & \underline{29.29} & 43.80 & 32.00 & 32.17 & 49.56 \\
\addlinespace[0.3em]
\midrule
\addlinespace[0.3em]
\rowcolor{turquoise!30} Luth-LFM2-350M & \textbf{57.05} & \textbf{28.28} & \underline{44.36} & \textbf{23.20} & \underline{34.81} & \underline{45.92} \\
SmolLM2-360M-Instruct & 33.95 & 20.71 & 26.18 & 3.00 & \textbf{35.41} & \textbf{52.17} \\
LFM2-350M & \underline{56.81} & \underline{27.27} & \textbf{44.79} & \underline{20.87} & 34.27 & 45.07 \\
\bottomrule
\end{tabular}
\caption{\label{tab:english-benches}
Results of Luth and other models on various English tasks. The scores are reported as percentages (Pass@1), averaged over three runs. The highest and second-best scores are shown in \textbf{bold} and \underline{underlined} respectively for each model category.
}
\end{table*}

\noindent \textbf{Main insights} \quad Luth models demonstrate that training on a high-quality, language-specific post-training dataset and leveraging model merging can lead to significant improvements in both French and English. Indeed, all Luth models substantially outperform their respective base models, as well as any model of comparable size, in French, while maintaining stable or even improved performance in English across widely used benchmarks. We attribute this phenomenon to cross-lingual transfer from French to English. Notably, Luth models exhibit average absolute score improvements in French ranging from \(+3.12\%\) to \(+11.26\%\) and in English from \(+0.76\%\) to \(+3.20\%\) across the six selected benchmarks. Furthermore, by fine-tuning the strongest SLMs available from two different families, we expect that our approach can substantially enhance the capabilities of any SLM under 2 billion parameters.

\section{Conclusion}

\noindent This paper introduces \textbf{Luth}, a family of state-of-the-art French SLMs that outperform all other models of comparable size on six French benchmarks covering general knowledge, instruction following, and mathematics. Although specialized in French, these models retain strong capabilities in other languages, particularly English, even showing improvements on various English benchmarks through cross-lingual transfer. These results stem from two key innovations: (1) the \textbf{Luth-SFT}, a French post-training dataset which drastically improves the model's performance in French and (2) \textbf{the use of model merging} to retain multilingual skills while further improving each component’s specialized language capabilities. Moreover, we demonstrate that careful fine-tuning on a specific language alone can yield significant performance gains without resorting to costly methods like continual pretraining. We expect that similar improvements could extend to larger architectures and other languages; verifying this remains a direction for future work.

\section{Limitations}

While Luth models achieve state-of-the-art performance, several limitations remain. First, our evaluation covers only a limited set of benchmarks; while they provide strong signals, they do not fully capture the models’ capabilities. 

Moreover, we assessed stability primarily in English without thoroughly evaluating whether the models retain their ability in other languages. Our experiments were also restricted to SLMs (under 2 billion parameters), which may limit the extent to which our approach unlocks potential gains at larger scales. 

Finally, the Luth-SFT dataset does not cover key capabilities such as tool use or code generation, which are increasingly central to modern LLMs.

\section*{Acknowledgments}
We thank Djamé Seddah and Thibaud Southiratn for comments on earlier versions of this work. This work was partly funded by the BPI Scribe project. 

\bibliography{custom}

\appendix

\twocolumn

\section{Luth-SFT System Prompts}
\label{sec:appendixA}

\subsection{Translation system prompt}
\label{sec:appendixTranslation}

\begin{tcolorbox}[
    colback=gray!10, 
    colframe=black, 
    width=\columnwidth, 
    boxrule=0.75pt, 
    left=4pt, right=4pt, top=4pt, bottom=4pt,
    breakable,
    enhanced jigsaw
]
\footnotesize
\textbf{System:} You are a professional French translator. Translate English text into natural, accurate French. \\

\textbf{REQUIREMENTS:}
\begin{itemize}[leftmargin=8pt, topsep=1pt, itemsep=0.5pt, parsep=0pt]
    \item Preserve exact meaning, tone, and register of the original
    \item Use natural French syntax and idiomatic expressions
    \item Maintain all formatting (markdown, HTML, special characters, structure)
    \item Keep technical terms, code snippets, and proper nouns appropriately handled
    \item Ensure grammatical correctness and contemporary French usage
\end{itemize}

\textbf{\\OUTPUT:}
\begin{itemize}[leftmargin=8pt, topsep=1pt, itemsep=0.5pt, parsep=0pt]
    \item Only the translated French text in identical format to the input.
\end{itemize}
\end{tcolorbox}

\subsection{General filtering system prompts}
\label{sec:appendixFiltering}

\begin{tcolorbox}[
    colback=gray!10, 
    colframe=black, 
    width=\columnwidth, 
    boxrule=0.75pt, 
    left=4pt, right=4pt, top=4pt, bottom=4pt,
    breakable,
    enhanced jigsaw
]
\footnotesize
\textbf{System:} You are a dataset quality assistant. Evaluate the question-answer pair below. \\

Return \texttt{False} if \emph{any} of the following apply:\\

\textbf{1. Code-Related}
\begin{itemize}[leftmargin=8pt, topsep=1pt, itemsep=0.5pt, parsep=0pt]
    \item Programming questions or answers
    \item Code snippets or syntax
    \item Mentions of languages, libraries, tools
    \item Debugging or optimization
\end{itemize}

\textbf{\\2. Tool Calling Content}
\begin{itemize}[leftmargin=8pt, topsep=1pt, itemsep=0.5pt, parsep=0pt]
    \item Describes or requests use of external tools, APIs, or systems
    \item Includes function calls, command-line usage, API requests, or tool invocation logic
    \item Involves configuring or troubleshooting external tools (e.g., databases, IDEs, browsers, CLIs)
\end{itemize}

\textbf{\\3. Logical Errors}
\begin{itemize}[leftmargin=8pt, topsep=1pt, itemsep=0.5pt, parsep=0pt]
    \item Contradictions, invalid reasoning, factual errors
\end{itemize}

\textbf{\\4. French Grammar Errors}
\begin{itemize}[leftmargin=8pt, topsep=1pt, itemsep=0.5pt, parsep=0pt]
    \item Wrong conjugations, tense, gender
    \item Bad structure, spelling, accents
\end{itemize}

\textbf{\\Respond only with:} \texttt{True} or \texttt{False}
\end{tcolorbox}

\begin{tcolorbox}[
    colback=gray!10, 
    colframe=black, 
    width=\columnwidth, 
    boxrule=0.75pt, 
    left=4pt, right=4pt, top=4pt, bottom=4pt,
    breakable,
    enhanced jigsaw
]
\footnotesize
\textbf{System:} Validate French Q\&A pairs.\\

Return True \textbf{only if} BOTH the question and the answer meet \textbf{all} of the following criteria:\\

\begin{itemize}[leftmargin=8pt, topsep=1pt, itemsep=0.5pt, parsep=0pt]
    \item Are written entirely in French.
    \item Are complete, grammatically correct, and coherent.
    \item Do not include any instruction to switch languages (e.g., ``answer in English'', ``répondez en anglais'', etc.).
    \item Do not contain mixed languages or foreign text (excluding proper nouns).
    \item The \textbf{question} must be an instruction or task prompt (e.g., ``Traduis ce texte...'',
          ``Explique...'').
    \item The \textbf{question} must \textbf{not} be a narrative, story, or purely informative content.
\end{itemize}

Return False if any of these conditions are not met and respond with \textbf{only}: \texttt{True} or \texttt{False}.
\end{tcolorbox}

\subsection{Scholar}
\label{sec:appendixScholar}

\subsubsection{Extraction of Question/Answer pairs system prompt}

\begin{tcolorbox}[
    colback=gray!10, 
    colframe=black, 
    width=\columnwidth, 
    boxrule=0.75pt, 
    left=4pt, right=4pt, top=4pt, bottom=4pt,
    breakable,
    enhanced jigsaw
]
\footnotesize
\textbf{System:} The following is the full text of a French high school exam:

\rule{\columnwidth}{0.5pt}

\texttt{\{subject\}}

\rule{\columnwidth}{0.5pt}

For each question below, extract its introductory description from the subject (such as problem description or setup). Do not summarize or rewrite. Return a JSON list of dictionaries with keys: \texttt{``question''} and \texttt{``context''}.\\

\textbf{Questions:}
\end{tcolorbox}

\subsubsection{Refinement and enrichment system prompt}

\begin{tcolorbox}[
    colback=gray!10, 
    colframe=black, 
    width=\columnwidth, 
    boxrule=0.75pt, 
    left=4pt, right=4pt, top=4pt, bottom=4pt,
    breakable,
    enhanced jigsaw
]
\footnotesize
\textbf{System:} You will receive three inputs: a question, a context, and an answer.\\

\textbf{REQUIREMENTS:}
\begin{itemize}[leftmargin=8pt, topsep=1pt, itemsep=0.5pt, parsep=0pt]
    \item Correct any errors in spelling, grammar, LaTeX, and formatting in all three inputs.
    \item Carefully review the context and correct it if there are any issues. If the context is missing or empty but should be present based on the question and answer, generate a relevant and useful context. But do not provide the answer to the question or a hint.
    \item Rephrase the answer to add clarity by:
    \begin{itemize}[leftmargin=8pt, topsep=0.5pt, itemsep=0.25pt, parsep=0pt]
        \item Expanding on the reasoning,
        \item Breaking the answer down into logical steps or explanations,
        \item Justifying the conclusion.
    \end{itemize}
    \item The final answer must remain logically and factually equivalent to the original.
    \item Do not change the overall intent of the question, context, or response unless necessary for correction.
    \item Provide your output strictly in JSON format with the following keys: \texttt{``question''}, \texttt{``context''}, and \texttt{``reponse''}.
    \item The entire output must be in French.
    \item Do not add any hints in the context related to the question.
\end{itemize}

\textbf{\\ENTRIES:}
\begin{itemize}[leftmargin=8pt, topsep=1pt, itemsep=0.5pt, parsep=0pt]
    \item \textbf{Question:} \texttt{\{question\}}
    \item \textbf{Context:} \texttt{\{context\}}
    \item \textbf{Response:} \texttt{\{response\}}
\end{itemize}
\end{tcolorbox}

\clearpage

\subsubsection{Example snippet from the dataset}

\begin{tcolorbox}[
    colback=gray!10, 
    colframe=black, 
    width=\columnwidth, 
    boxrule=0.75pt, 
    left=4pt, right=4pt, top=4pt, bottom=4pt,
    breakable,
    enhanced jigsaw
]
\footnotesize
\vspace{0.3cm}

\textbf{Role: User}
\begin{quote}
\textit{La suite $(u_n)$ est définie pour tout entier naturel $n$ par $u_n = \int_0^1 x^n \ln(1+x) \mathrm{d}x$.\\
Calculer $u_0$ à l'aide d'une intégration par parties.}
\end{quote}

\vspace{0.3cm}

\textbf{Role: Assistant}
\begin{quote}
\textit{Pour calculer $u_0$, nous utilisons la définition de la suite $(u_n)$ pour $n=0$.}

\textbf{Étape 1 : Expression de $u_0$}\\
\textit{En remplaçant $n$ par 0 dans l'expression générale, nous obtenons :}\\
$u_0 = \int_0^1 x^0 \ln(1+x) \mathrm{d}x = \int_0^1 \ln(1+x) \mathrm{d}x$.

\textbf{Étape 2 : Application de l'intégration par parties}\\
\textit{Nous calculons cette intégrale en utilisant la formule d'intégration par parties :}\\
$\int_a^b f(x)g'(x) \mathrm{d}x = [f(x)g(x)]_a^b - \int_a^b f'(x)g(x) \mathrm{d}x$.

\textit{Pour notre intégrale $\int_0^1 \ln(1+x) \mathrm{d}x$, nous pouvons considérer que $\ln(1+x)$ est multiplié par 1. Nous posons alors :}
\begin{itemize}
\item $f(x) = \ln(1+x)$, \textit{ce qui donne} $f'(x) = \frac{1}{1+x}$.
\item $g'(x) = 1$. \textit{Pour la primitive $g(x)$, nous choisissons astucieusement $g(x) = x+1$.}
\end{itemize}

\textbf{Étape 3 : Calcul de l'intégrale}\\
\textit{En appliquant la formule avec nos fonctions, on obtient :}\\
$u_0 = [\ln(1+x)(x+1)]_0^1 - \int_0^1 \frac{1}{1+x}(x+1) \mathrm{d}x$.

\textit{Simplifions et calculons chaque terme séparément :}
\begin{enumerate}
\item \textbf{Le terme entre crochets :}\\
$[(x+1)\ln(1+x)]_0^1 = 2\ln(2) - 0 = 2\ln(2)$
\item \textbf{L'intégrale restante :}\\
$\int_0^1 \frac{1}{1+x}(x+1) \mathrm{d}x = \int_0^1 1 \mathrm{d}x = 1$
\end{enumerate}

\textbf{Étape 4 : Conclusion}\\
$u_0 = 2\ln(2) - 1$.
\end{quote}
\end{tcolorbox}

\clearpage

\section{Training details}
\label{sec:appendixB}

\begin{table}[htbp]
\centering
\small
\caption{Hyperparameters used to train \textbf{Luth-0.6B-Instruct} (Qwen3-0.6B) on a single Nvidia H100 80GB RAM.}
\begin{tabular}{l c}
\toprule
\textbf{Hyperparameter} & \textbf{Value} \\
\midrule
Learning rate           & $2 \times 10^{-5}$ \\
Batch size (per device) & 6 \\
Gradient accumulation   & 4 \\
Optimizer               & AdamW (8-bit) \\
Weight decay            & 0.01 \\
Gradient clipping       & 0.1 \\
Warmup steps            & 264 \\
Scheduler               & Cosine \\
Max sequence length     & 16,384 \\
Training epochs         & 3 \\
Max training steps      & 2640 \\
Precision               & bfloat16 \\
Gradient checkpointing  & True \\
Flash Attention         & True \\
Packing                 & True \\
\bottomrule
\end{tabular}
\label{tab:hyperparams_luth_0.6b}
\end{table}

\begin{table}[htbp]
\centering
\small
\caption{Hyperparameters used to train \textbf{Luth-1.7B-Instruct} (Qwen3-1.7B) on a single Nvidia H100 80GB RAM.}
\begin{tabular}{l c}
\toprule
\textbf{Hyperparameter} & \textbf{Value} \\
\midrule
Learning rate           & $2 \times 10^{-5}$ \\
Batch size (per device) & 3 \\
Gradient accumulation   & 8 \\
Optimizer               & AdamW (8-bit) \\
Weight decay            & 0.01 \\
Gradient clipping       & 0.1 \\
Warmup steps            & 264 \\
Scheduler               & Cosine \\
Max sequence length     & 16,384 \\
Training epochs         & 3 \\
Max training steps      & 2640 \\
Precision               & bfloat16 \\
Gradient checkpointing  & True \\
Flash Attention         & True \\
Packing                 & True \\
\bottomrule
\end{tabular}
\label{tab:hyperparams_luth_1.7b}
\end{table}

\begin{table}[htbp]
\centering
\small
\caption{Hyperparameters used to train \textbf{Luth-LFM2-350M} (LFM2-350M) on a single Nvidia H100 80GB RAM.}
\begin{tabular}{l c}
\toprule
\textbf{Hyperparameter} & \textbf{Value} \\
\midrule
Learning rate           & $5 \times 10^{-5}$ \\
Batch size (per device) & 8 \\
Gradient accumulation   & 2 \\
Optimizer               & AdamW (torch\_fused) \\
Weight decay            & 0 \\
Gradient clipping       & 0.1 \\
Warmup steps            & 407 \\
Scheduler               & Cosine \\
Max sequence length     & 16,384 \\
Training epochs         & 3 \\
Max training steps      & 4074 \\
Precision               & bfloat16 \\
Gradient checkpointing  & True \\
Flash Attention         & True \\
Packing                 & True \\
\bottomrule
\end{tabular}
\label{tab:hyperparams_luth_lfm2_350m}
\end{table}

\begin{table}[htbp]
\centering
\small
\caption{Hyperparameters used to train \textbf{Luth-LFM2-700M} (LFM2-700M) on a single Nvidia H100 80GB RAM.}
\begin{tabular}{l c}
\toprule
\textbf{Hyperparameter} & \textbf{Value} \\
\midrule
Learning rate           & $5 \times 10^{-5}$ \\
Batch size (per device) & 12 \\
Gradient accumulation   & 3 \\
Optimizer               & AdamW (torch\_fused) \\
Weight decay            & 0.01 \\
Gradient clipping       & 0.1 \\
Warmup steps            & 270 \\
Scheduler               & Cosine \\
Max sequence length     & 16,384 \\
Training epochs         & 3 \\
Max training steps      & 2709 \\
Precision               & bfloat16 \\
Gradient checkpointing  & True \\
Flash Attention         & True \\
Packing                 & True \\
\bottomrule
\end{tabular}
\label{tab:hyperparams_luth_lfm2_700m}
\end{table}

\begin{table}[htbp]
\centering
\small
\caption{Hyperparameters used to train \textbf{Luth-LFM2-1.2B} (LFM2-1.2B) on a single Nvidia H100 80GB RAM.}
\begin{tabular}{l c}
\toprule
\textbf{Hyperparameter} & \textbf{Value} \\
\midrule
Learning rate           & $4 \times 10^{-5}$ \\
Batch size (per device) & 8 \\
Gradient accumulation   & 4 \\
Optimizer               & AdamW (torch\_fused) \\
Weight decay            & 0 \\
Gradient clipping       & 0.1 \\
Warmup steps            & 203 \\
Scheduler               & Cosine \\
Max sequence length     & 16,384 \\
Training epochs         & 3 \\
Max training steps      & 2037 \\
Precision               & bfloat16 \\
Gradient checkpointing  & True \\
Flash Attention         & True \\
Packing                 & True \\
\bottomrule
\end{tabular}
\label{tab:hyperparams_luth_lfm2_1.2b}
\end{table}

\end{document}